\documentclass[conference]{IEEEtran}
\IEEEoverridecommandlockouts
\usepackage{cite}
\usepackage{amsmath,amssymb,amsfonts}
\usepackage{subcaption}
\usepackage{algorithmic}
\usepackage{graphicx}
\usepackage{textcomp}
\usepackage{xcolor}
\usepackage{cancel}
\usepackage{booktabs}
\usepackage{balance}
\usepackage{fancyhdr}

\newcommand{\fs}[1]{{\color{cyan}{#1}}}

\def\BibTeX{{\rm B\kern-.05em{\sc i\kern-.025em b}\kern-.08em
    T\kern-.1667em\lower.7ex\hbox{E}\kern-.125emX}}
\begin{document}

\title{Design of a Real-time Asynchronous %/Event-based 
Monocular Odometry for Planetary Exploration %\\
}

\author{\IEEEauthorblockN{Be\~nat I\~nigo\IEEEauthorrefmark{1}\IEEEauthorrefmark{2}, Florian Steidle\IEEEauthorrefmark{1}, Wolfgang St{\"u}rzl\IEEEauthorrefmark{1}}
\IEEEauthorblockA{\IEEEauthorrefmark{1}\textit{Institute of Robotics and Mechatronics} \\
\textit{German Aerospace Center (DLR)}\\
%Germany \\
%email address or ORCID
}
\IEEEauthorblockA{\IEEEauthorrefmark{2}
\textit{University of Zaragoza} %\\
%\textit{German Aerospace Center (DLR)}
}

%\and
}

\maketitle
\fancypagestyle{withfooter}{
\renewcommand{\headrulewidth}{0pt}
\fancyfoot[C]{\footnotesize Accepted to the Challenges and Opportunities of Neuromorphic Field Robotics and Automation IEEE ICRA Workshop - 2026}
}
\thispagestyle{withfooter}
\pagestyle{withfooter}
\begin{abstract}
We describe our preliminary design of a real-time asynchronous event-based monocular odometry for planetary exploration. Operating under strict computational constraints, planetary rovers frequently encounter complex, unpredictable environments that demand high-speed sensing and robustness to high dynamic range (HDR) lighting. Event cameras address these needs by reporting asynchronous, pixel-wise brightness changes with microsecond resolution, significantly reducing data bandwidth while maintaining robustness in extreme lighting conditions. We propose an approach based on an Error-State Kalman Filter (ESKF) that leverages this asynchronous event stream to continuously estimate camera ego-motion. The camera state is updated with every tracked position output generated by RATE, a real-time asynchronous feature tracker.

%\rem{*CRITICAL: Do Not Use Symbols, Special Characters, Footnotes, 
%or Math in Paper Title or Abstract.}
\end{abstract}

%\begin{IEEEkeywords}
%component, formatting, style, %styling, insert
%\end{IEEEkeywords}

\section{Introduction}
\begin{figure}[htbp]
\centering
%\centerline{\includegraphics{fig1.png}}
\includegraphics[height=2.85cm]{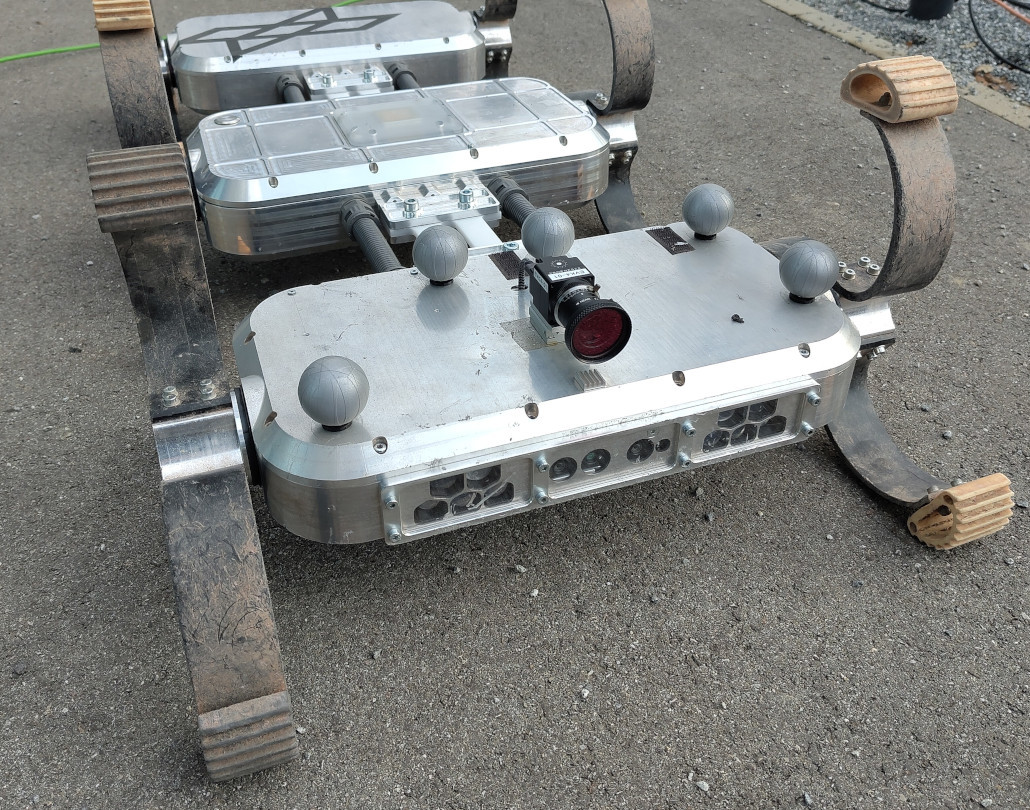}
\includegraphics[height=2.85cm]{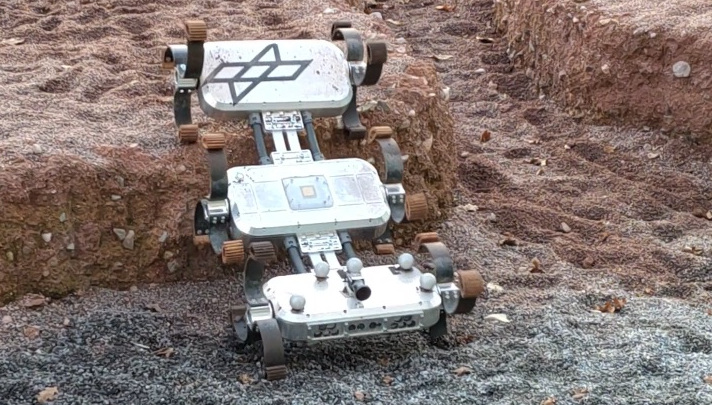}
\caption{Scout rover with event camera mounted on top its head segment}
\label{fig:scout_with_eventcam}
\end{figure}

Due to their low power consumption and data bandwidth requirements, as well as  high dynamic range and low latency, event cameras have great potential for planetary exploration robotics. Having conducted field tests at the DLR Moon Mars test site~\cite{MoonMarsTestSite} with the Scout rover, see Fig.~\ref{fig:scout_with_eventcam}, we decided to employ event cameras for pose estimation. 
In the following  we describe our preliminary design of an event-based monocular visual odometry 
with the focus on low latency pose estimation on standard off-the-shelf  CPUs.
Note, that although fusing data from different sensors, e.g. camera and inertial measurement unit (IMU), is usually more robust, there are also space rovers that predominantly rely on vision, like the MMX rover IDEFIX  that due to size constraints and low gravity on Mars Moon Phobos does not use inertial measurements for state estimation~\cite{NavDLR}.

\section{Related Work}

Pure event-driven monocular visual odometry remains sparsely explored, with only few published methods that function independently of both frame data and inertial sensors. 
One of the first approaches~\cite{Kim2016} employs three interleaved filters for motion, intensity and depth estimation.
EVO~\cite{Rebecq2017} which uses parallel tracking and mapping, does not require intensity estimation, allowing it to run faster.
Pose tracking of the camera is realized using Lukas-Kanade-based alignment of a template created by projecting the 3D map and an event image created by accumulating events within an short time span. To initialize the system, the local scene is assumed to be planar and fronto-parallel to the sensor. 
An asynchronous optimization-based method is described in~\cite{Liu2022}. A Gaussian Process (GP) is used to model the trajectory of the camera. Similar to our approach HASTE is employed for feature tracking. However, the implementation of the full pipeline did not run in real time. 
In~\cite{Wang2025}, a sliding-window optimizer for GP regression is introduced to reduce the computational complexity. While the event-based feature tracking seems to be similar to our frontend, we employ an Error-State Kalman Filter (ESKF) for online estimation of camera poses, as described in the next section.
Recent learning-based approaches, e.g. DEVO~\cite{Klenk2024}, can estimate camera trajectories with high accuracy. However, they need powerful GPUs currently not available on planetary rovers.

%In~\cite{Mourikis2007} the authors introduced the Multi-State Constraint Kalman Filter (MSCKF).
%This filter does not include the 3D feature positions in the state. 
%The measurement equation enforces multi-view geometric constraints across several camera poses and removes the dependency of the measurement residual on the feature position using nullspace projection.
%This keeps the computational complexity of the filter linear in the number of features while still using rich visual information.
%The benefits of this approach come with some weaknesses.
%E.g. feature tracks are not used immediatly, but a delayed update strategy is used.
%In~\cite{Zhang2025}, the authors proped an immediate update strategy. 
%It imporves accuracy and consistency of the filtering results.
%The core idea is to use feature observations immediately after image processing is finished.
%Therefore, a single feature track produces multiple sequential updates and improves linearization quality as states are corrected earlier.
%In~\cite{du2024po} ... \\
%Is this the correct paper~\cite{Diel2005} to cite for epipolar constraint filtering? \\
%In this paper, we propose to ... \\
%The main contributions are
%\begin{enumerate}
%    \item 1
%    \item 2
%\end{enumerate}

\section{Monocular Odometry with ESKF}
For detection, tracking and management of feature\fs{s} we employ RATE~\cite{RATE}.
To facilitate real-time tracking of multiple event-based features, RATE distributes features into sub-images. Features are detected with Shi-Tomasi corner detection~\cite{ShiTomasi1994} on a binarized Surface of Active Events~\cite{Benosman2014,Yilmaz2021} and then tracked asynchronously with HASTE~\cite{HASTE}. Changes of feature states (position $\pm 1$\,px, orientation $\pm 4^\circ$) are reported via ROS messages. 

We added a second output to RATE, a list of IDs for features to be deleted. Features are removed when they leave the image boundaries, when their tracking quality degrades, or when a higher-scoring feature is detected within the same sub-region. 
\subsection{Error-State Kalman Filter (ESKF)}
Messages received from the frontend are used to update the state
%(equations et al.)
%The robot state can be represented as
%In our approach the state is 
defined as
\begin{equation}
    x = \begin{bmatrix} ^{G}p_{C}^\top & ^{G}q_{C}^\top & ^{G}v_{C}^\top & ^{G}p_{f_{1}}^\top & \dots & ^{G}p_{f_{n}}^\top \end{bmatrix}^\top %\enspace,
    \label{eq:nominal_state}
\end{equation}
where $^{G}p_{C}$, $^{G}q_{C}$ and $^{G}v_{C}$ represent the position, orientation and velocity of the camera in the global frame, and ${}^{G}p_{f_{1}} \dots {}^{G}p_{f_{n}}$ represent the set of all landmarks reconstructed by triangulating features from corresponding observations.  As new track updates are received from RATE composed as $h_j =    (id_j,t_j, (u_j,v_j))$, i.e. feature ID, timestamp, and new feature position, we can update the %robot's 
state and possibly triangulate new features. 

We adopt an error state formulation of the Kalman filter, defining the the discrepancy between the true state $x_{\rm true}$ and the nominal state as 
\begin{equation}
    \delta x = \begin{bmatrix} ^{G}\delta p_{C}^\top & ^{G}\delta \theta_{C}^\top& ^{G}\delta v_{C}^\top & ^{G}\delta p_{f_{1}}^\top & \dots & ^{G}\delta p_{f_{n}}^\top \end{bmatrix}^\top
    \label{eq:error_state}
\end{equation}
such that $x_{\rm true} = x \oplus \delta x$. The  orientation error $^{G}\delta \theta_{C}$ is defined as a 3D rotation vector as a minimal representation. We represent the uncertainty of the system via the covariance matrix $P$:
\begin{equation}
    P = \begin{bmatrix}
          P_{cc} & P_{cf} \\ P_{fc} & P_{ff}
    \end{bmatrix}
\end{equation}
where $P_{cc}$ represents the camera pose and velocity covariance, $P_{ff}$ the map features covariance, and $P_{cf}$ the cross-covariance between them.
\subsubsection{Propagation}
We propagate the state forward in time $\delta t$ assuming a constant velocity and constant orientation kinematic model. The prediction step of the filter is given by
\begin{equation}
\begin{aligned}
    \hat{x}_{k|k-1} &= f(\hat{x}_{k-1|k-1}, u_{k-1}) \\ 
    P_{k|k-1} &= F_kP_{k-1|k-1}F_k^\top + Q_{k-1}
\end{aligned}
\end{equation}
where $P_{k|k-1} \in \Re^{(9+3M)\times (9+3M)}$ is the error state covariance for $M$ landmarks, $Q_{k-1} \in \Re ^{(9+3M)\times (9+3M)}$ the process uncertainty and $F_k \in \Re ^{(9+3M)\times (9+3M)}$ the Jacobian of the error-state kinematics evaluated at the zero-mean error state:
\begin{equation}
    F_k = \left. \frac{\partial f}{\partial \delta x_{k-1}} \right|_{\delta x_{k-1} =0}
\end{equation}
\subsubsection{Correction}
Every time we receive a feature update, we check whether it corresponds to an existing feature in the state. If the feature is new, we clone the initial state for further initialization, as detailed in Section~\ref{sec:initialization}.

If it is an existing feature, we compute the innovation $\tilde{y}_k$ defined as the difference between the actual observation $z_k$ and the predicted measurement $\hat{z}_k$:
\begin{equation}
    \tilde{y}_k = z_k - h(\hat{x}_{k|k-1}) = z_k - \hat{z}_k
\label{eq:innovation}
\end{equation}

We present the measurement model  $h(\cdot)$ by considering an update of a single feature track $h_j$, observed from pose $({}^{G}p_{C_j} , {}^{G}q_{C_j})$. We project 3D feature $j$ into the camera's image plane:
\begin{equation}
    \hat{z}_{j} = \begin{bmatrix} \hat{u}_{j} \\ \hat{v}_{j} \end{bmatrix} = \begin{bmatrix} f_x \frac{^{C}X_j}{^{C}Z_j} + c_x \\ f_y \frac{^{C}Y_j}{^{C}Z_j} + c_y \end{bmatrix}
\label{eq:projection_model}
\end{equation}
%\ws{[any reason not to use lower index $j$, ie. $\hat{z}_j$, $\hat{u}_j$, $\hat{v}_j$?]}
by first transforming the global feature position ${}^{G}p_{f_{j}}$ into the camera frame $C$:
\begin{equation}
    ^{C}p_{f_{j}} = \begin{bmatrix} {}^{C}X_j \\ ^{C}Y_j \\ ^{C}Z_j \end{bmatrix} = R(^{G}q_{C})^\top \left( ^{G}p_{f_{j}} - {}^{G}p_{C} \right)
\label{eq:camera_frame_transformation}
\end{equation}
where $^{G}p_{f_{j}}$ is the 3D feature position in the camera frame. Note that we receive already undistorted pixel coordinates from RATE via a Look-Up Table (LUT). \\
As measurements are asynchronous, we process one measurement at a time, defining a sparse measurement Jacobian matrix $H_k \in \Re^{2 \times (9+3M)}$ mapping to the state vector:
\begin{equation}
\begin{aligned}
    H_k = \begin{bmatrix} H_{pos} & H_{rot} & 0_{2 \times 3} & 0_{2 \times 3} & \dots & H_{f_j} & \dots & 0_{2 \times 3} \end{bmatrix}
\end{aligned}
\label{eq:full_H_jacobian}
\end{equation}
where $H_{pos}$ and $H_{rot} \in \Re ^{2 \times 3}$ are the Jacobians of the pixel error with respect to the camera position and orientation, and $H_{f_j} \in \Re ^{2 \times 3}$ is the Jacobian with respect to the 3D landmark, evaluated at the zero error state:
\begin{equation}
\begin{aligned}
    H_{pos} &= \left. \frac{\partial z}{\partial ^{G} \delta p_{C}} \right|_{\delta x_{k-1} =0} \\
    H_{rot} &= \left. \frac{\partial z}{\partial ^{G}\delta\theta_{C}} \right|_{\delta x_{k-1} =0} \\
    H_{f_i} &= \left. \frac{\partial z}{\partial ^{G} \delta p_{f_{j}}} \right|_{\delta x_{k-1} =0}
\end{aligned}
\label{eq:H_jacobians}
\end{equation}
We compute the Kalman gain $K_k \in \Re^{(9+3M)\times 2}$ and update the state and covariance:
\begin{equation}
\begin{aligned}
    S_k &= H_kP_{k|k-1}H_k^\top+R_k \\
    K_t &= P_{k|k-1}H_k^\top S_k^{-1} \\
    \delta \hat{x}_{k|k} &= K_k\tilde{y}_k \\
    \hat{x}_{k|k} &= \hat{x}_{k|k-1} \oplus \delta \hat{x}_{k|k} \\
    P_{k|k} &= (I-K_kH_k)P_{k|k-1} \\
\end{aligned}
\label{eq:kalman_update}
\end{equation}
 where $\oplus$ defines the composition between the estimated nominal state and the error state. 
 \subsubsection{Reset}
Finally, we apply the ESKF error reset operation, defined by %ing the error reset 
function $g(\cdot)$ as:
\begin{equation}
\begin{aligned}
    \delta x \leftarrow g(\delta x) = \delta x \ominus \delta \hat{x}_{k|k-1}
\end{aligned}
\label{eq:reset_function}
\end{equation}
where $\ominus$ stands for the composition inverse of $\oplus$. The reset operation on the state and covariance is thus:

\begin{equation}
\begin{aligned}
    \delta \hat{x}_{k|k} &\leftarrow 0 \\
    P_{k|k} &\leftarrow GP_{k|k}G^\top
\end{aligned}
\label{eq:reset_operation}
\end{equation}
with $G$ denoting the Jacobian of the reset operation:
\begin{equation}
G = \left. \frac{\partial g}{\partial \delta x} \right|_{\delta x_k = \delta \hat{x}_{k|k}}
\label{eq:reset_jacobian}
\end{equation}
 
\subsection{3D Feature/Landmark Initialization}
\label{sec:initialization}
If the feature track $h_j = (id_j,t_j,(u_j,v_j))$ does not correspond to any existing landmark in the state, we must save the camera pose from which this new feature was observed for the first time for future triangulation. 
Because this past pose depends on future corrections, we save it by augmenting the state using a stochastic cloning update~\cite{1014801}. We delay triangulation until sufficient parallax is achieved, at which point we augment the state with the newly triangulated feature. 

Considering a single measurement $z_k$, the respective augmented state and covariance matrix become:
\begin{equation}
\begin{aligned}
    \delta \bar{x} &= C 
    %\delta x = \begin{bmatrix}
    %      I \\ I^*
    %\end{bmatrix} %_{(9+3N+6)\times(9+3N)}
    \begin{bmatrix}
          \delta x_k
    \end{bmatrix}_{(9+3M)} =
    \begin{bmatrix}
          \delta x_k \\ \delta x_{clone}
    \end{bmatrix}_{(9+3M+6)} \\
    \bar{P} &= C P_{k|k} C^\top\\
    C & = \begin{bmatrix}
          I \\ I^*
    \end{bmatrix} _{(9+3M+6)\times(9+3M)}
\end{aligned}
\end{equation}
where $I^*$ is a sparse selection matrix that extracts the 6-DoF camera pose (position and orientation) from the current state to be cloned. 

To ensure real-time performance, we marginalize out lost features from the state vector by removing its corresponding rows and columns from the state vector and the covariance matrix. %Furthermore, because the features tracked by RATE \ws{are spatio-temporal rather than purely geometric [meaning unclear to me]}, a track is permanently discarded once it moves out of the frame. If the same physical feature is observed again, it is initialized as a distinct new landmark. 

Once sufficient parallax is established between the cloned state and the current nominal state, we triangulate the new feature position and add it to the state. As the total number of landmarks is not known a priori, the state vector expands incrementally. We can define the state expansion function $q(\cdot)$ for feature $i$ as:
\begin{equation}
\begin{aligned}
    \bar{x}_k &= q(x_{k-1}, z_{k},  z_{k-j}, ^{G}p_{C_{k-1}}, ^{G}p_{C_{k-j}}) \\
    &= \begin{bmatrix}
          x_{k-1} \\ g(^{G}p_{C_{k-1}}, ^{G}q_{C_{k-1}}, ^{G}p_{C_{k-j}},^{G}q_{C_{k-j}}, z_k, z_{k-j})
    \end{bmatrix} \\
    &= \begin{bmatrix}
                x_{k-1} \\ ^{G}p_{f_{i_{k-1}}}
          \end{bmatrix}
\end{aligned}
\end{equation}
where $^{G}p_{C_{k-j}}$ represents the camera pose cloned for this specific feature, and $g(\cdot)$ is the %non-linear algebraic inverse-depth 
triangulation 
function that computes the feature's position $^{G}p_{f_i}$.

After a new landmark $j$ is triangulated, the expanded state vector becomes:
\begin{equation}
    \bar{x}_{k+1} \leftarrow \begin{bmatrix}
          x_k & ^{G}p_{f_j}
    \end{bmatrix}^\top 
\end{equation}
By linearizing the triangulation function, we estimate the error of the 3D landmark position:
\begin{equation}
    ^{G} \delta p_{f_{j}} \approx G_x \delta x_k + G_z \delta z_k
\end{equation}
%\ws{[$^{G}\delta  p_{f_{i}}$ instead of $\delta ^{G} p_{f_{i}}$..?]}
where $G_x$ is the Jacobian of the triangulation with respect to the augmented state vector and $G_z$ is the Jacobian with respect to the measurement noise. Due to the high non-linearity of the inverse-depth formulation, $G_x$ and $G_z$ are evaluated numerically via finite differences:
\begin{equation}
    G_x \approx \frac{\Delta g_k}{\Delta x_k},\ G_z \approx \frac{\Delta g_k}{\Delta z_k}
\end{equation}
%\ws{[maybe $G_x \approx \frac{\Delta g_k}{\Delta x_k},\ G_z \approx \frac{\Delta g_k}{\Delta z_k}$ instead]}
We then augment the covariance matrix accordingly:

\begin{equation}
    \bar{P} =\begin{bmatrix}
          P_{xx}&P_{xm}\\
          P_{mx}&P_{mm}
    \end{bmatrix} = \begin{bmatrix}
          P & PG_x^\top \\
          G_xP & G_xPG_x^\top+G_zRG_z^\top
    \end{bmatrix}
\end{equation}

\subsection{Filter Initialization using Homography}
\begin{figure}[htbp] % [htbp] is generally safer than [H] in two-column layouts
    \centering
    % Primera imagen
    \begin{subfigure}{0.75\columnwidth} % Changed to \columnwidth
        \centering
        \includegraphics[width=\linewidth]{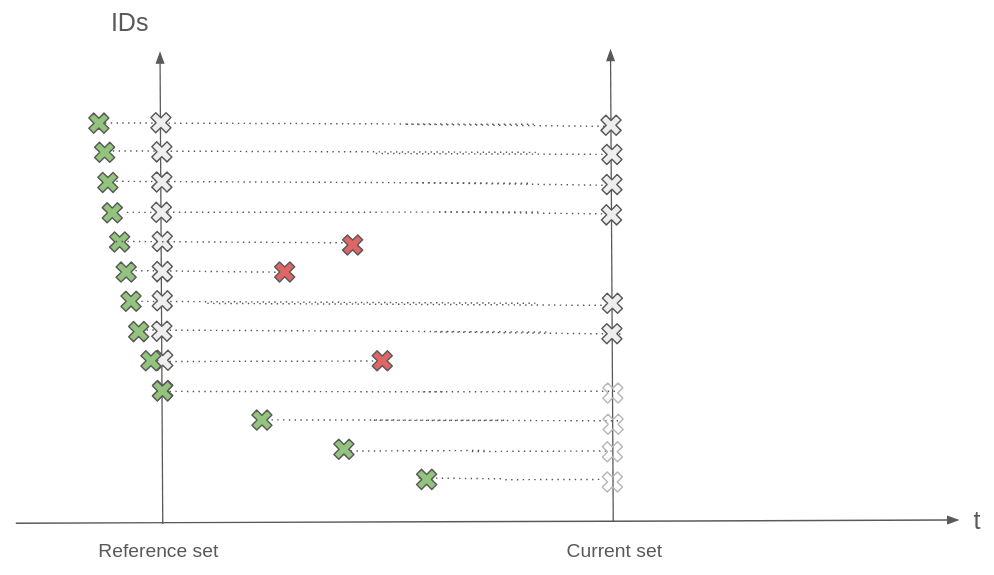} 
    \end{subfigure}
    \caption{Event sets for initial triangulation, green showing birth and red showing termination of tracks. Vertical axis for IDs and horizontal axis for time.}
    \label{fig:initialization}
\end{figure}

Our current ESKF-based approach assumes that the 3D position of at least a small number of features are known or can be triangulated. Therefore we need a separate procedure to bootstrap the system at the start when neither 3D landmarks nor camera poses are known, which is based on a classical computer vision method.
Because state updates of features are not synchronized like those from conventional frame-based cameras, we group them into sets, see Fig.~\ref{fig:initialization}. We wait until the first $N$ feature updates arrive, which are then used as a reference set. The current set is initialized as a clone of the reference and updated every time a track update from RATE arrives. In order to make the tracks more stable, we apply a Kalman filter to each individual feature to act as a smoother. Once sufficient parallax is achieved, we initialize the system, i.e. estimate the motion between the current and the reference set using homography and then triangulate feature 3D positions. Since we currently rely on homography for this setup, we choose a planar dataset.

\section{Preliminary Results}
To validate the proposed Error-State Kalman Filter (ESKF) pipeline for asynchronous event-based tracking, we conducted preliminary evaluations in both a controlled simulation environment and on a real-world event camera dataset.
\subsection{Simulation Evaluation}
To rigorously evaluate the filter's behavior, we developed a custom simulation environment. The simulator generates spatially distributed 3D landmarks that dynamically appear and disappear, mimicking the realistic scenario of physical features being initialized and subsequently lost. The tracking front-end processes asynchronous events generated from a simulated sensor. To %accurately 
emulate the asynchronous nature of event cameras and the continuous feature updates from the RATE tracker, the active 3D landmarks are projected onto the 2D image plane sequentially at random order. Furthermore, to evaluate the system's robustness against the measurement noise inherent to real event sensors, we introduced a synthetic reprojection error by injecting zero-mean Gaussian noise $\mathcal{N}(0, \sigma^2)$ to the 2D pixel coordinates. 

Because the system operates as a purely monocular pipeline without IMU integration, state propagation relies strictly on a constant-velocity kinematic model. To account for unmodeled dynamics when undergoing sharp accelerations, we empirically tune the process noise covariance as $\sigma_a, \sigma_w = 2.0$, and the measurement noise is set to $\sigma_u=\sigma_v = 1.0$ pixels.

\begin{figure}[htbp] % [htbp] is generally safer than [H] in two-column layouts
    \centering
    % Primera imagen
    \begin{subfigure}{\columnwidth} % Changed to \columnwidth
        \centering
        \includegraphics[width=\linewidth]{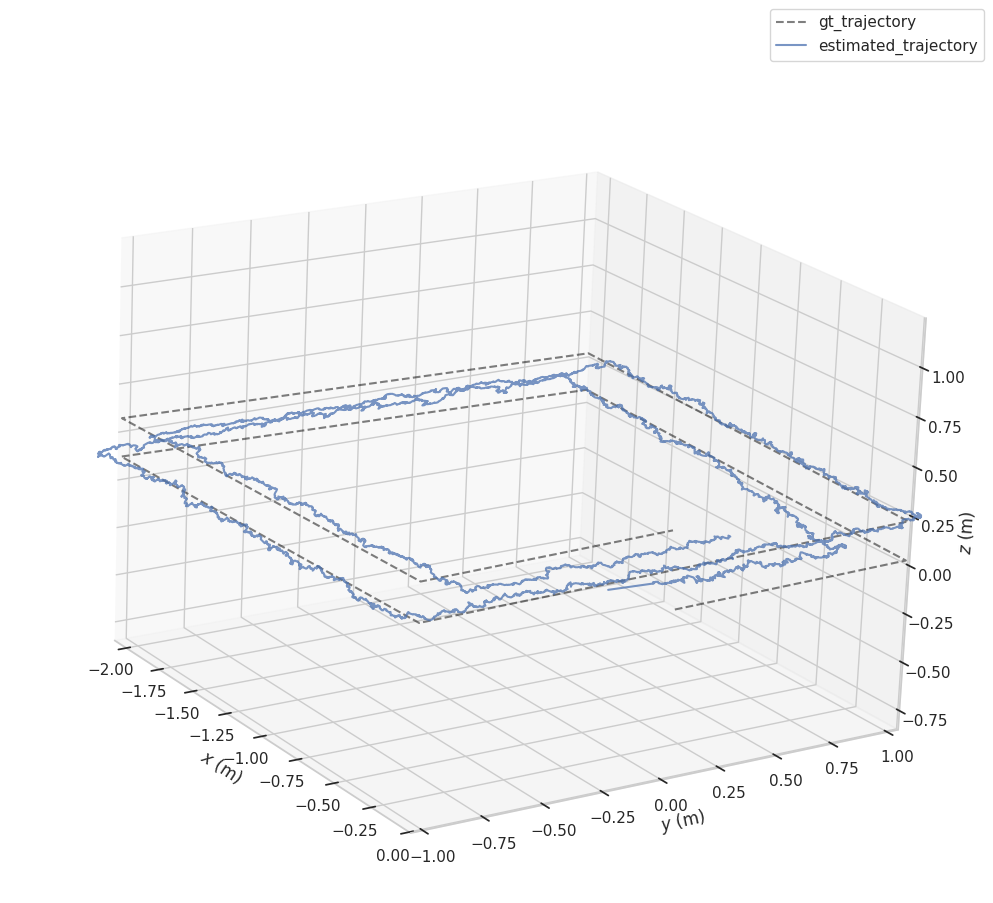} 
        \caption{Robot trajectory}
    \end{subfigure}
    
    \vspace{0.5cm} % Vertical gap
    
    % Segunda imagen
    \begin{subfigure}{\columnwidth} % Changed to \columnwidth
        \centering
        \includegraphics[width=\linewidth]{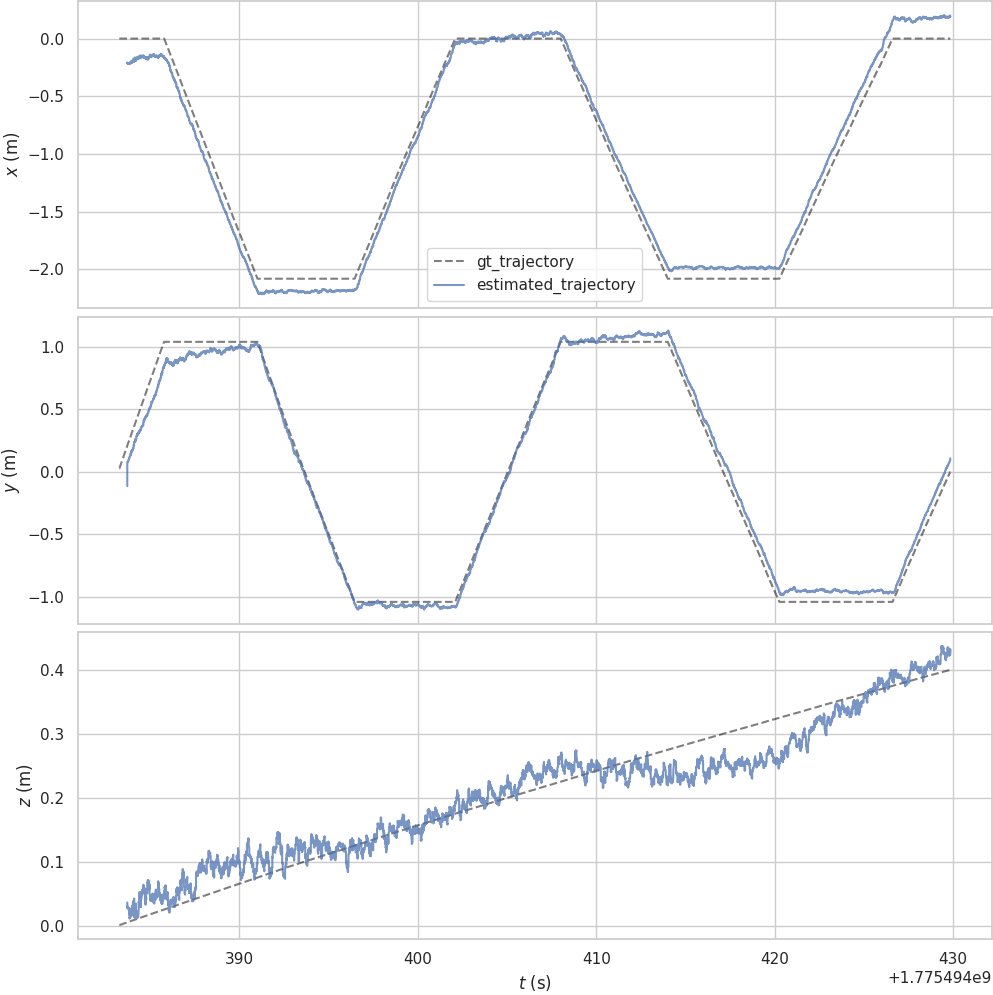} 
        \caption{Trajectory error}
    \end{subfigure}
    
    \caption{Plot shows the estimated trajectory against the ground truth for noise level $\sigma=1.0$ pixel. For convenience, \cite{grupp2017evo} was used for evaluation. After alignment mean absolute pose error is 0.065\,m.}
    \label{fig:sim_traj}
\end{figure}

Due to the lack of metric scale observability, the estimated trajectory was aligned to the ground truth using a $\text{Sim}(3)$ Umeyama alignment. Preliminary simulation results validate the core asynchronous update formulation, see Fig.~\ref{fig:sim_traj} for an example. The high-frequency variance observed in the estimated trajectory is characteristic of heuristically tuned noise covariance matrices, where the measurement noise covariance $R_k$ currently outweighs the process noise covariance $Q_k$, leading to over-correction from noisy pixel observations.

%\begin{table}[htbp]
%    \centering
%    \caption{Absolute Pose Error metrics in simulation.}
%    \label{tab:sim_ape_metrics}
%    \begin{tabular}{lc}
%        \toprule
%        \textbf{Metric} & \textbf{Absolute Pose Error (m)} \\
%        \midrule
%        RMSE & 0.126 \\
%        Mean & 0.115 \\
%        Median & 0.108 \\
%        Std. Dev. & 0.051 \\
 %       Max & 0.391 \\
 %       Min & 0.001 \\
%        \bottomrule
%    \end{tabular}
%\end{table}

\subsection{Real-World Evaluation}
For real-world validation, we evaluated our system using the widely benchmarked UZH DAVIS Event Camera Dataset \cite{mueggler2017event}. Specifically, initial testing was conducted on the wall poster sequence.

\begin{figure}[htbp] % [htbp] is generally safer than [H] in two-column layouts
    \centering
    % Primera imagen
    \begin{subfigure}{\columnwidth} % Changed to \columnwidth
        \centering
        \includegraphics[width=\linewidth]{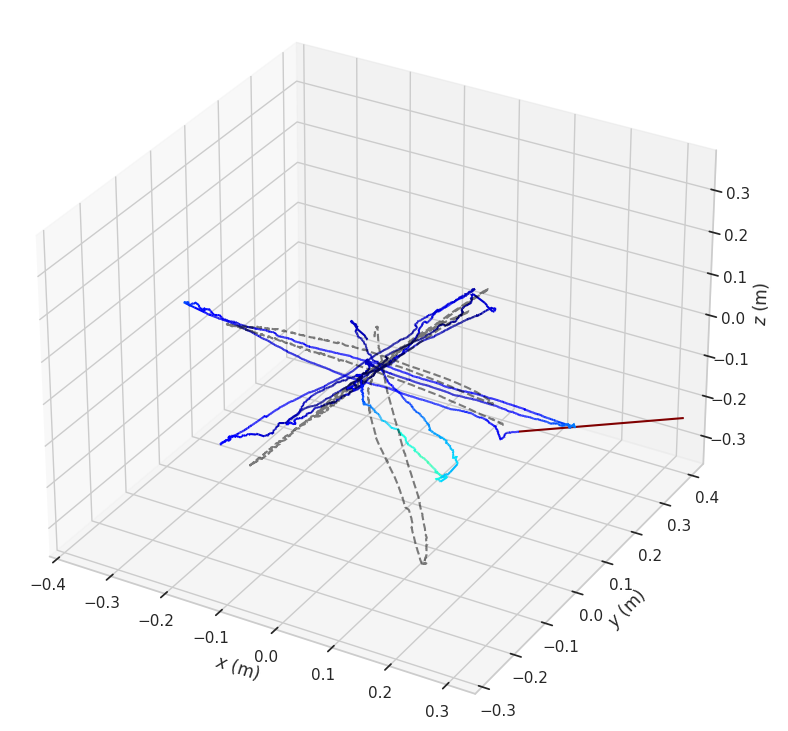} 
        \caption{Robot trajectory}
    \end{subfigure}
    
    \vspace{0.5cm} % Vertical gap
    
    % Segunda imagen
    \begin{subfigure}{\columnwidth} % Changed to \columnwidth
        \centering
        \includegraphics[width=\linewidth]{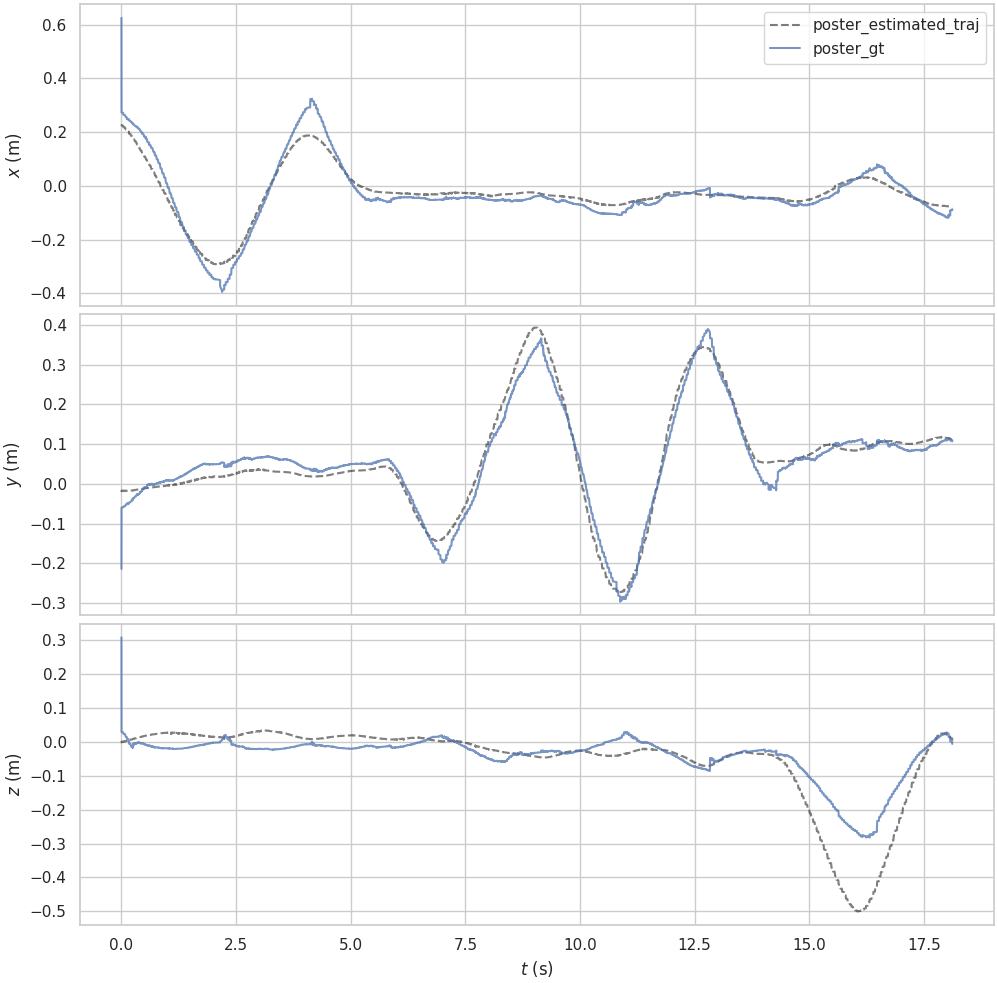} 
        \caption{Trajectory error}
    \end{subfigure}
    
    \caption{Plot shows the estimated trajectory against ground truth. After alignment mean absolute pose error is 0.06\,m. Total trajectory length is approx. 6.5\,m. The red line indicates initialization with homography.}
    \label{fig:poster_traj}
\end{figure}

Following with the simulation methodology, the estimated trajectory was aligned to the ground truth using a $\text{Sim}(3)$ Umeyama alignment. 
%Due to the non-deterministic nature of the RATE outputs, 
A representative  example is shown in Fig.~\ref{fig:poster_traj}. After successful initialization, the camera pose could be tracked with an RMS of 0.06\,m. After about 17.5\,s pose estimation fails due to loss of the majority of feature tracks.
%This demonstrates the theoretical potential of the proposed formulation, though future work will focus on stabilizing run-to-run variance.

%\begin{table}[htbp]
%    \centering
%    \caption{Absolute Pose Error metrics in UZH DAVIS Event Camera Dataset.}
%    \label{tab:ape_metrics}
%    \begin{tabular}{lc}
%        \toprule
%        \textbf{Metric} & \textbf{Absolute Pose Error (m)} \\
%        \midrule
%        RMSE & 0.067 \\
%        Mean & 0.055 \\
%        Median & 0.049 \\
%        Std. Dev. & 0.037 \\
%        Max & 0.236 \\
%        Min & 0.006 \\
%        \bottomrule
%    \end{tabular}
%\end{table}

\section{ %Discussion and
Next Steps %% / Future Work
} 
There are at least two aspects of  the described approach that we want to improve in the near future.
 \paragraph{Improved pose estimation during feature initialization} 
 In time intervals where the number of features in the filter state is low due to tracking losses, accuracy of the estimated camera poses can drop significantly. In these cases of tracking loss or feature deletion, RATE attempts to detect and track new features. However, even if successful, features cannot be triangulated without sufficient parallax and thus camera pose estimation cannot be enhanced immediately. To improve the situation and make earlier use of recently detected and tracked features, we propose to add a measurement error based on epipolar constraint, which does not depend on features being triangulated.    
\paragraph{Improved filter initialization}
For non-planar scenes, the current initialization based on homography has to be replaced with a more general approach, e.g. using 5pt-algorithm~\cite{Nister5pt} in a RANSAC-based approach~\cite{RANSAC}. 
However, a filter-based initialization, possibly using epipolar residuals, is preferred as it better aligns with the asynchronous stream of feature state updates. 

We are currently trying to address (a) and (b) with a modified approach %based on inverse depth parameterization 
which will be tested and evaluated with respect to run-time, latency and accuracy against the described method and existing baseline approaches.
Preliminary investigations show that CPU load is dominated by the frontend, i.e. feature tracking and management.  
\paragraph*{Scale ambiguity}
While it is interesting to investigate what can be achieved with a single event camera, it is also clear that our monocular VO can estimate camera poses only up to an unknown scale unless additional information is given. In a planetary scenario this could be, for example, the lander, a reference tag of known size on the lander or an already mapped crater. 
Also, additional sensors that can provide scale information, like a laser range finder or an IMU, could be utilized.
%\vspace*{6cm}
%\addtolength{\textheight}{-12cm}
%Do not use \verb|\nonumber| inside the \verb|{array}| environment. It
%will not stop equation numbers inside \verb|{array}| (there won't be
%any anyway) and it might stop a wanted equation number in the
%surrounding equation.

\balance
%\section*{Acknowledgment?}

\bibliographystyle{IEEEtran}
\bibliography{literature}

\end{document}